\documentclass{llncs}
\usepackage{graphicx}
\usepackage{url}

\usepackage{amsmath}
\usepackage{amsfonts}
\usepackage{amsthm}
\usepackage{amssymb}
\usepackage{bm}
\usepackage{bbm}
\usepackage{xcolor}
\usepackage{booktabs}
\usepackage{comment}
\usepackage{caption}

\graphicspath{{./imgs/}}

\title{On the Validity of Bayesian Neural Networks for Uncertainty Estimation}
 \author{John Mitros and Brian Mac Namee}
 \institute{School of Computer Science\\ University College Dublin, Dublin, IR \\
 \email{\{ioannis,\,brian.macnamee\}\,@\,insight-centre.ie} }

\begin{document}
\maketitle

\begin{abstract}
  Deep neural networks (DNN) are versatile parametric models utilised successfully in a diverse number of tasks and domains. However, they have limitations---particularly from their lack of robustness and over-sensitivity to out of distribution samples. Bayesian Neural Networks, due to their formulation under the Bayesian framework, provide a principled approach to building neural networks that address these limitations. This paper describes a study that empirically evaluates and compares Bayesian Neural Networks to their equivalent point estimate Deep Neural Networks to quantify the predictive uncertainty induced by their parameters, as well as their performance in view of this uncertainty. In this study, we evaluated and compared three point estimate deep neural networks against comparable Bayesian neural network alternatives using well-known benchmark image classification datasets.
\end{abstract}

\keywordname{ Bayesian Neural Networks, Uncertainty Quantification, OoD, Robustness}

\section{Introduction}
\label{sec:intro}
With the advancement of technology and the abundance of data, our society has been transformed beyond recognition. From smart home assistance technologies to self-driving cars, to smart mobile phones a multitude of connected devices now assist us in our daily routines.

One thing that is common among these devices is the exponential explosion of data generated as a consequence of our activities. Predictive models rely on this data  to capture patterns in our daily routines, from which they can offer us assistance tailored to our individual needs. Many of these predictive models are based on deep neural networks (DNNs).

The machine learning community, however, is becoming increasingly aware of issues associated with DNNs ranging from fairness to bias, and, from robustness to uncertainty estimation. Motivated by this we have investigated the issues of reliability and trustworthiness of prediction confidence estimates produced by DNNs.
We first assess the capability of current DNN models to provide confident (i.e.~calibration error) and reliable (i.e.~noise sensitivity error or the ability to predict out of sample instances with high uncertainty) predictions. Second, we compare this  to the capability of equivalent recent Bayesian formulations (i.e.~Bayesian neural networks (BNN)), in terms of accuracy, calibration error, and ability to recognise and indicate out of sample instances.

There exist two types of uncertainties related to predictive models, \emph{aleatoric} and \emph{epistemic} uncertainty~\cite{gal2015}. Aleatoric uncertainty is usually attributed to stochasticity inherent in the related task or experiment to be performed. It can therefore be considered an irreducible error. Epistemic uncertainty is usually attributed to uncertainty induced by the model parameters. It can therefore be considered a a reducible error as it can be reduced by obtaining more data. The question under investigation in this work is whether BNNs can provide better calibrated and reliable estimates for out of sample instances compared to point estimate DNNs and therefore relates to epistemic uncertainty.

The remaining sections of this paper are divided as follows. Section \ref{sec:related_work} outlines related work in the areas of prediction confidences and Bayesian neural networks. In Section~\ref{sec:data}, we provide the information related to the datasets used throughout the experiments, their respective sizes and types. In Section~\ref{sec:metrics}, we describe the metrics used to evaluate whether a classifier is calibrated (i.e.~expected calibration error and reliability figures), as well as its ability to identify and predict out of sample instances (i.e.~symmetric KL divergence and distributional entropy figures), along with their respective explanations. In Section~\ref{sec:methods}, we introduce the three BNN approaches utilised in the experiments, providing detailed explanations of how they work. Finally, in Sections~\ref{sec:results} and~\ref{sec:conclusion} we present the results related to confidence calibration (i.e.~Table~\ref{tab:ece} and Figure~\ref{fig:reliability}) and reliability prediction estimates for out of sample instances (i.e.~Table~\ref{tab:divergence} and Figures~\ref{fig:entropy_cifar},~\ref{fig:entropy_svhn},~\ref{fig:entropy_fmnist}), along with the concluding remarks.

\section{Related Work}
\label{sec:related_work}
Earlier findings~\cite{nalisnick2019,schulam2019,stracuzzi2018,shafaei2018,choi2018} have demonstrated the incapacity of point estimate deep neural networks (DNN) to provide confident and calibrated~\cite{lee2017a,guo2017,kumar2019} uncertainty estimates in their predictions. Motivated by recent work in this area we strive to demonstrate, first, that this is indeed a serious problem currently investigated in the machine learning community, and second, provide an alternative viable solutions (i.e.~BNN) which combines the best from both worlds (i.e.~a principled and elegant formulation due to the Bayesian inference framework and powerful and expressive models thanks to DNN).

In order to help the reader understand the terminology and semantics of uncertainty quantification in predictive models, it would be helpful to relate the variance existent in the model parameters represented as the sum of both aleatoric and epistemic uncertainty. Additionally, whenever the reader encounters the following terminology ``point estimate DNN" in the document, it simply refers to a DNN model coupled with a softmax function in the final layer. For instance, suppose our DNN is defined by $\hat{\mathbf{y}} = f(\mathbf{x})$, where $\hat{\mathbf{y}}$ denotes the predictions for $K$ possible classes. Then a ``point estimate DNN" is simply defined by exponentiating each prediction and normalising it by the total sum of all exponentiated predictions. 
\begin{align}
	 p(y=j\vert\mathbf{x}) &= \frac{e^{\hat{\mathbf{y}_{i}}}}{\sum_{j=1}^{K} e^{\hat{\mathbf{y}_{j}}}}\;\mbox{for}\, i=1,\ldots,K\;\mbox{and}\;\hat{\mathbf{y}} = (y_{1},\ldots,y_{K})\in\mathbb{R}^{K}
~\label{eq:softmax}
\end{align}

The reason for identifying them as point estimate DNN is because they are mistakenly misinterpreted as probabilistic models due to the fact that they provide predictions which resemble probabilities (i.e.~estimates $\in\mathbb{R}_{[0-1]}$). Furthermore, Eq.~\ref{eq:softmax} is misinterpreted mistakenly for a categorical distribution to which we disagree since it should have a prior Dirichlet in order to be classified as a categorical distribution.  Our view is that Eq.~\ref{eq:softmax} is more of a mathematical convenience in order to allow DNN model to emit predictions resembling predictions rather than a well defined probability distribution.

As previously stated there are two main problems investigated in this work. The first one is related to the inability of DNN to predict probability estimates representative of the true correct likelihood function (i.e.~calibration confidence). For instance, in a binary classification task a classifier is considered uncalibrated when the classifiers' predictions do not match the empirical proportion of the positive class upon which the classifier is requested to make a prediction. Poor calibration confidence problems in DNN can be affected by different choices while constructing the DNN architecture such as for instance the depth, width, regularisation or batch-normalisation~\cite{guo2017,kumar2019}. The second problem, is related to the incapacity of DNN to identify and reliably predict out of sample instances (i.e.~noise sensitivity) which can be a consequent of noise in the data, noise in the model parameters or noise constructed by an adversary in order to manipulate the models' predictions~\cite{choi2018,maddox2019,shafaei2018,fawzi2018}.

\section{Data}
\label{sec:data}
The overall data used in this empirical study include three well-established datasets in the machine learning literature, \textit{CIFAR-10}, \textit{SVHN}, and \textit{FashionMNIST}. The first two datasets are comprised of colour images of dimensionality 32x32 and include 10 distinct categories while the last one contains grayscale images of dimensionality 28x28. In addition, the first two are considered to represent real world datasets with \textit{CIFAR-10} being collected over the Internet while \textit{SVHN} being a result of the Google Street View project representing house numbers. Further details regarding the number of instances of each dataset and equivalently their categories are described below
\begin{itemize}
    \item The \textit{CIFAR-10} dataset consists of 60,000 colour images of dimensionality 32x32 with 10 classes. Each class contains 6,000 images. In total there are 50,000 training images and 10,000 test images.
    
    \item The \textit{SVHN} dataset consists of 99,289 colour images of dimensionality 32x32 representing digits of house numbers. There exist 10 categories one for each digit, in total there are 73,257 colour images representing digits for training, and equivalently 26,032 digits for testing.
    
    \item The \textit{FahsionMNIST} dataset consists of 60,000 examples and a test set of 10,000 examples. Each example is a 28x28 grayscale image, associated with a label from 10 classes.
\end{itemize}

\section{Metrics}
\label{sec:metrics}
The chosen evaluation metrics utilised for this empirical study involved:
\begin{itemize}
    \item Accuracy
    \item Expected calibration error
    \item Entropy
    \item Symmetric $D_{KL}$ divergence
\end{itemize}

Particularly, for a given neural network model $\hat{y} = f(\mathbf{x};\bm{\theta})$ of depth $L$ defined as $\{\mathbf{W}_{L}\sigma_{L}(\mathbf{W}_{L-1}\ldots\sigma_{2}(\mathbf{W}_{2}\sigma_{1}(\mathbf{W}_{1}\mathbf{x})))\}$, describing the number of function compositions and parameters $\{\bm{\theta} = \{\mathbf{W}_{1},\ldots,\mathbf{W}_{L}\}$ with $\sigma(\cdot)$ being a nonlinear function.

The accuracy on a given output $\hat{y}_{n}$ is measured by the indicator function $\mbox{acc} = \frac{1}{N}\sum_{n=1}^{N}\mathbbm{1}(y_{n}\neq\hat{y}_{n})$ for each instance $n\in N$ averaged over the total number of instances $N$ in the dataset. This metric is predominantly used in the machine learning community to evaluate the generalisation ability of a predictive model on a hold-out test set.

In order to capture whether a model is calibrated we utilised the expected calibration error $ECE = \sum_{m=1}^{M}\vert\mbox{acc}(B_{m}) - \mbox{conf}(B_{m})\vert\frac{\vert B_{m}\vert}{N}$ in combination with the equivalent reliability plots shown in Figure~\ref{fig:reliability} similar to~\cite{guo2017}. ECE is usually expressed as a weighted average between the accuracy and confidence of a model across $M$ bins for $N$ samples. This metric has the ability to capture any disagreement between the classifiers predictions and the true empirical proportion of instances for each class category for every mini-batch of instances presented to the classifier.

Furthermore, to assess a models' ability to characterise out of sample data with high degree of uncertainty we focused on the work of~\cite{maddox2019} utilising information entropy $H(Y)=-\sum_{k=1}^{K} p(\hat{\mathbf{y}}_{k})\log p(\hat{\mathbf{y}}_{k})$ on the final predictions of a model in order to derive the uncertainty plots depicted in Figure~\ref{fig:entropy_cifar}. 
Essentially, for every input $\mathbf{x}$ we have we have a corresponding vector of predictions $\hat{\mathbf{y}} = (0.86, 0.23,\ldots, K)$ where each entry denotes the prediction of the classifier for each class $k\in K$.  For every dataset we split them randomly into two halves. The first half represents the $K/2$ classes and the other other half the remaining. We select one of the halves to be utilised to train the classifier (i.e.~denotes in-sample instances) and the remaining half (i.e.~denotes out-of-sample instances) to be utilised only during the testing phase of the classifier. Therefore, after the classifier has been trained on one half (hence the 5+5 categories in Figures~\ref{fig:entropy_cifar}
) we evaluate its generalisation ability on the remaining half where for every input we have a corresponding entropy over the $K$ classes. This provides a distribution over the total number of $N$ inputs allowing to distinguish and evaluate the classifier entropy among in-sample vs out-of-sample instances.  

Finally, in order to conveniently compare and summarize a models' performance on detecting out of sample instances as a summary statistic the scalar value of the symmetric KL-divergence $D_{KL}(p\parallel q) + D_{KL}(q\parallel p)$~\cite{maddox2019} between two distributions $p$ and $q$ was selected as a sensible candidate. The choice of KL-divergence allows to evaluate how similar are two distributions $p$ and $q$. The larger the KL-divergence is the more distinct are the distributions $p$ and $q$. Since we want to evaluate the ability of the classifier to recognise out of sample instances we should be able to measure the KL-divergence of the classifiers' estimates for in-sample $p$ against out of sample $q$ instances. The larger KL values denote the  classifier is in better position to recognise out of sample instances.

\section{Methods}
\label{sec:methods}
This section provides the details of the empirical evaluation comprised of the following three components, (i) models, (ii) calibration and (iii) uncertainty. The following (i) models were selected among which three of them represent point estimate deep neural networks (DNN) and the remaining three their equivalent Bayesian Neural Networks (BNN).
\begin{itemize}
    \item Point estimate deep neural networks
    \begin{itemize}
        \item VGG16~\cite{simonyan2015}
        \item PreResNet164~\cite{he2016}
        \item WideResnet28x10~\cite{zagoruyko2017}
    \end{itemize}
    \item Bayesian neural networks
    \begin{itemize}
        \item VGG16 - Monte Carlo Dropout~\cite{gal2015}
        \item VGG16 - Stochastic Weight Aaveraging of Gaussian samples~\cite{maddox2019}
        \item Stochastic Variational - Deep Kernel Learning~\cite{wilson2016}
    \end{itemize}
\end{itemize}

(ii) The calibration of each model was evaluated using the expected calibration error introduced in Section~\ref{sec:metrics} in combination with the reliability plots demonstrated in Figure~\ref{fig:reliability}. Each model was trained on 5 categories from \textit{CIFAR-10} and accordingly \textit{SVHN} with the remaining 5 categories being whithheld in order to evaluate the models' ability to associate out of samples instances with high uncertainty as they were not introduced to the model at any step. The duration of training for each model was 300 epochs with best performing model on the validation set being selected as the final model for each architecture.

(iii) As already stated in order to evaluate a model's ability to detect out of sample instances with high uncertainty we utilised entropy on the predictions of a model to derive Figures~\ref{fig:entropy_cifar},~\ref{fig:entropy_svhn} and~\ref{fig:entropy_fmnist}
for each dataset and model combination accordingly. In addition, the symmetric KL-divergence described in Section~\ref{sec:metrics} was introduced in order to provide a comparable scalar summary statistic of the overall essence of Figures~\ref{fig:entropy_cifar},~\ref{fig:entropy_svhn} and~\ref{fig:entropy_fmnist}. 

\begin{figure}[ht!]
    \centering
    \includegraphics[width=3in]{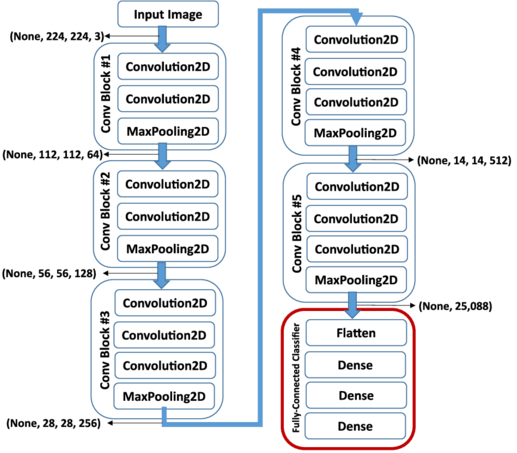}
    \caption{VGG16 model architecture~\cite{simonyan2015}.}
    \label{fig:vgg16}
\end{figure}

\begin{figure}[ht!]
    \centering
    \includegraphics[width=3in]{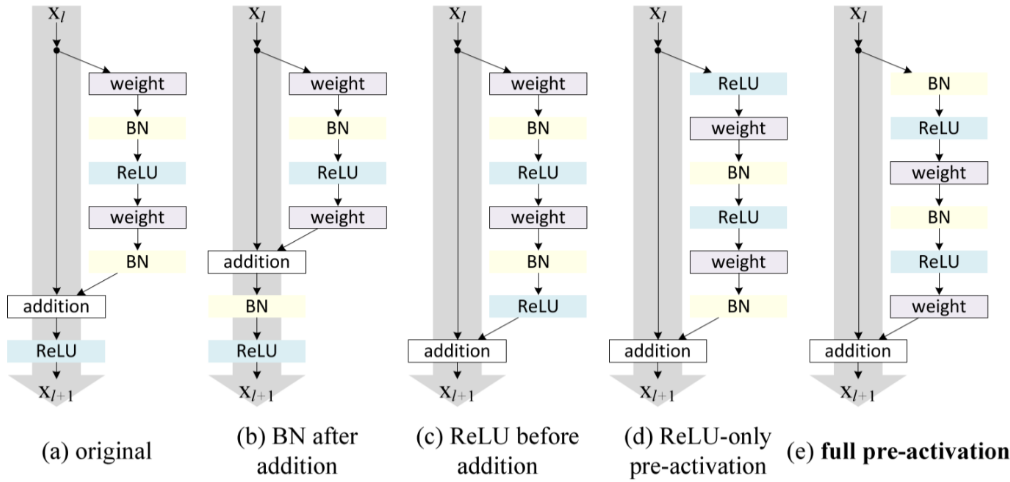}
    \caption{Various components of ResNet~\cite{he2016}.}
    \label{fig:preres}
\end{figure}

\begin{figure}[ht!]
    \centering
    \includegraphics[width=3in]{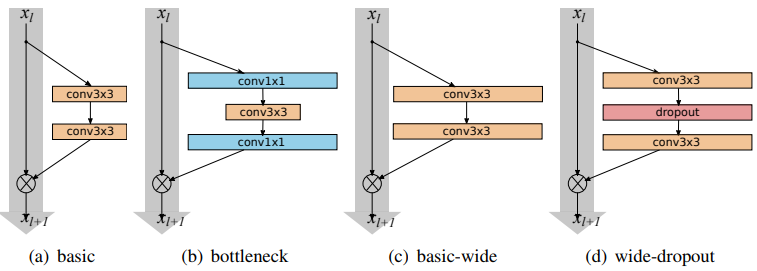}
    \caption{Various components of WideResNet~\cite{zagoruyko2017}.}
    \label{fig:wideres}
\end{figure}

In the remainder of this section we will introduce the three Bayesian neural network approaches utilised during the experimental study:
\begin{enumerate}
    \item Dropout as Bayesian approximation:
    
     This method~\cite{gal2015} provides a view of dropout at test time as approximate Bayesian inference. It is based on prior work of~\cite{damianou2013} which established a relationship between neural networks with dropout and Gaussian Processes (GP). Given a dataset $(\textbf{X}, \textbf{Y})$ the posterior over the GP is formulated as,
     \begin{align*}
        &\textbf{F}\vert\textbf{X}\sim\mathcal{N}(0, \textbf{K}(\textbf{X}, \textbf{X})) \\
        & \textbf{Y}\vert\textbf{F}\sim\mathcal{N}(\textbf{F}, 0\cdot\textbf{I}) \\
        & \hat{y}\vert\textbf{Y}\sim\mbox{Categorical}\left(\cdot\right)
     \end{align*}
    where $\hat{y}$ denotes a class label and $\hat{y}\neq\hat{y}\prime$. An integral part of the GP is the choice of the covariance matrix $\textbf{K}$ representing the similarity between two inputs as a scalar value. The key insight to draw connections between neural networks and Gaussian processes is to consider the possibility the choice of the kernel to represent a non-linear function, for instance, consider the rectified linear (ReLU) function, then the kernel would be expressed as $\int p(\textbf{w})\sigma(\textbf{w}^{T}\textbf{x})\sigma(\textbf{w}^{T}\textbf{x})d\textbf{w}$ with $p(\textbf{w})\sim\mathcal{N}\left(\bm{\mu}, \mathbf{\Sigma}\right)$. Because usually the integral is intractable a conventional approach would be to use Monte Carlo integration in order to approximate it $\hat{k} = \frac{1}{T}\sum_{t=1}^{T}\sigma(\textbf{w}_{t}^{T}\textbf{x})\sigma(\textbf{w}_{t}^{T}\textbf{x})$, hence, the name Monte Carlo Dropout. Let us now consider a one hidden layer neural network with dropout $\hat{y} = (\beta_{2}\mathbf{W}_{2})\sigma(\mathbf{x}(\beta_{1}\mathbf{W}_{1}))$ where $\beta_{1},\beta_{2}\sim\mbox{Bernoulli}(p_{1,2})$.
    Utilising the approximate kernel $\hat{k}$
    one can express the parameters $\mathbf{W}_{1,2}$ as $\mathbf{W}_{1,2} = \beta_{1,2}(\mathbf{A}_{1,2} + \bm{\sigma}\bm{\epsilon}_{1,2})(1 - \beta_{1,2})\bm{\sigma}\bm{\epsilon}_{1,2}$ with $\mathbf{A}_{1,2}, \bm{\epsilon}_{1,2}\sim\mathcal{N}(\mathbf{0}, \mathbf{I})$ and $\beta_{1,2}\sim\mbox{Bernouli}(p_{1,2})$ closely resembling the NN formulation.
    Therefore, to establish the final connection among NNs trained with stochastic gradient descent (SGD) and dropout to GPs one has to simulate Monte Carlo sampling by drawing samples from the trained model at test time $\left[\hat{y}_{n} = f(\mathbf{x}_{n}; \beta_{n}\bm{\theta}_{n})\right]_{n=1}^{N}$ with $\beta_{n}\sim\mbox{Bernouli}(p_{n})$. The samples $\hat{y}_{n}$ resulting from the different dropout masks $\beta_{n}$ are averaged over the $N$ different models in order to approximate and retrieve the posterior distribution. \\
    
    \item Stochastic weight averaging of Gaussian samples
    
    Stochastic weight averaging of Gaussian samples (SWAG)~\cite{maddox2019} is an extension of stochastic weight averaging (SWA)~\cite{izmailov2018} where the weights of a NN are averaged during different SGD iterates, which in itself can be viewed as approximate Bayesian inference~\cite{blei2017}, with ideas traced back to~\cite{ruppert1988,polyak1992}. In order to understand SWAG we first need to explain SWA. SWA at a high level can be viewed as averaged SGD~\cite{ruppert1988,polyak1992}. Essentially the main difference between SWA and averaged SGD is that SWA utilises a simple moving average instead of an exponential one, in conjunction with a high constant learning rate, instead of a decaying one. An illustration of the internal mechanics of SWA is demonstrated in Figure~\ref{fig:swa}. In essence, in SWA one maintains a running average over the weights of a NN during the last 25\% of the training process which is used to update the first and second moments of batch-normalisation. This leads to better generalisation since the SGD projections are smoothed out during the average process leading to wider optima in the optimisation landscape of the NN. Now that we have established what SWA is let us introduce SWAG. SWAG is an approximate Bayesian inference technique for estimating the covariance from the weight parameters of a NN.
    SWAG maintains a running average $\bar{\theta^{2}} = \frac{1}{T}\sum_{t=1}^{T}\theta_{t}^{2}$ in order to compute the covariance $\bm{\Sigma} = \mbox{diag}(\Bar{\theta^{2}} - \theta^{2})$ which produces the approximate Gaussian posterior $\mathcal{N}(\theta, \bm{\Sigma})$. At test time the weights of the NN are drawn from this posterior $\tilde{\theta_{n}}\sim\mathcal{N}(\theta, \bm{\Sigma})$ in order to
    perform Bayesian model averaging to retrieve the final posterior of the model as well as the uncertainty estimates from the first and second moments.
    
    \begin{figure}[ht!]
        \centering
        \includegraphics[width=3.5in]{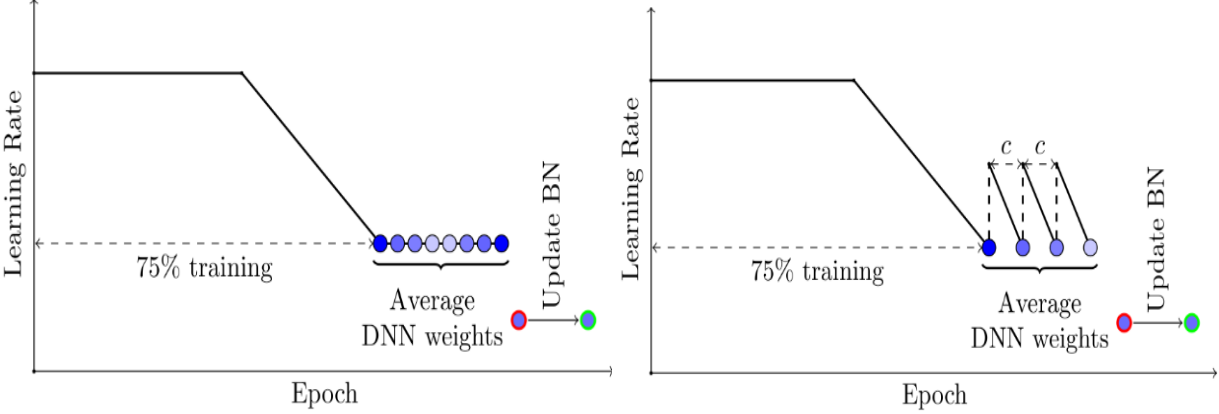}
        \caption{Illustration of the internal mechanics of SWA with different learning rate approaches. In both cases the SWA averages are formed during the last 25\% of training~\cite{izmailov2018}.}
        \label{fig:swa}
    \end{figure}
     
    \item Deep kernel learning
    
    The deep kernel learning method~\cite{wilson2016} establishes a combination of NN architectures and GPs trained jointly in order to derive kernels with GP properties overcoming the need to perform approximate Bayesian inference.
    The first part is composed of any NN (i.e.~task dependent) whose output is utilised in the second part in order to approximate the covariance of the GP in the additive layer
    As explained earlier in the MC-Dropout approach a kernel between inputs $\mathbf{x}$ and $\mathbf{x}\prime$ can be expressed via a non-linear mapping function thanks to the kernel trick $k(\mathbf{x}, \mathbf{x}\prime)\rightarrow k(f(\mathbf{x}, \mathbf{w}), f(\mathbf{x}\prime, \mathbf{w})\vert\mathbf{w})$ therefore combining NNs with GPs seems like a natural evolution which permits scalable and flexible kernels represented as neural networks to be utilised directly in Gaussian Processes. Finally, given that the formulation of GPs allows to represent a distribution over a function space it is thus possible to derive uncertainty estimates from the moments of this distribution in order to inform the models about the uncertainty in their parameters having an impact in the final posterior distribution.
  
    \begin{figure}[ht!]
        \centering
        \includegraphics[width=3in]{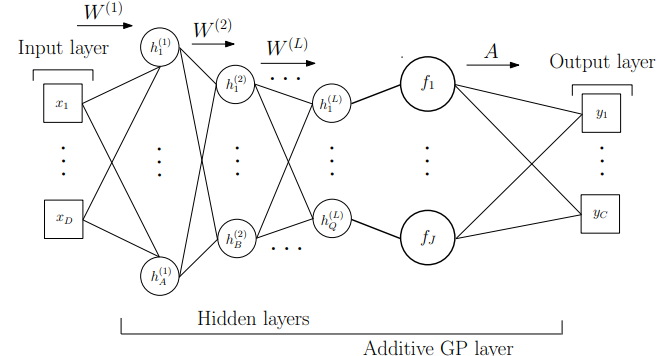}
        \caption{Stochastic variational deep kernel learning (i.e.~deep Gaussian processes)~\cite{wilson2016}.}
        \label{fig:deepgp}
    \end{figure}
\end{enumerate}

\section{Results}
\label{sec:results}
In this section we describe the results from our experiments and the findings that arise from these for our two initial questions. Let us recall them here again for clarity:
\begin{itemize}
    \item Do point estimate deep neural networks suffer from pathologies of poor calibration and inability to identify out of sample instances?
    \item Are Bayesian neural networks better calibrated and more resilient to out of sample instances?
\end{itemize}

\begin{figure}[ht!]
    \centering
    \includegraphics[width=4.8in]{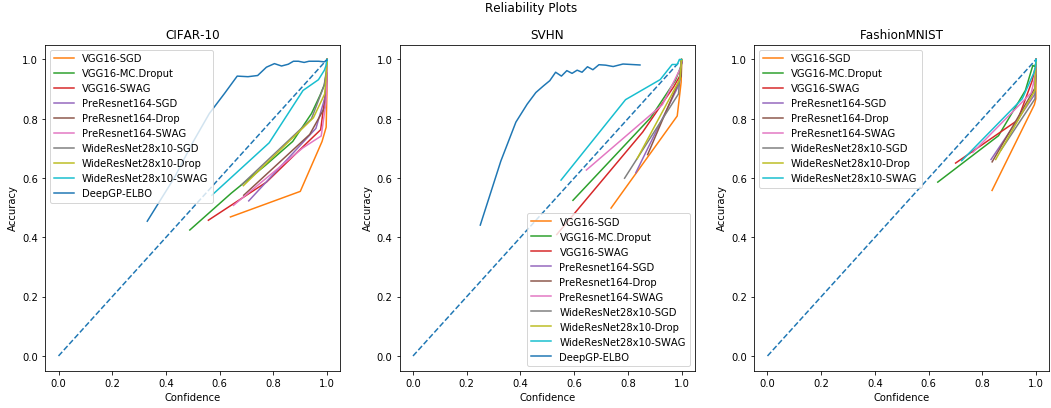}
    \caption{Reliability plots across all models on \textit{CIFAR-10}, \textit{SVHN} and \textit{FashionMNIST} datasets.}
    \label{fig:reliability}
\end{figure}

To answer the first question we  draw the attention of the reader to Figure~\ref{fig:reliability} and  equivalently Table~\ref{tab:ece}. Figure~\ref{fig:reliability} shows the reliability plots for all models and datasets. In these plots a perfectly calibrated model is indicated by the diagonal line. Anything below the diagonal represents an over-confident model, while anything above the diagonal represents an under-confident model. The expected calibration errors (ECE) in Table~\ref{tab:ece} (which measure the degree of miscalibration present) seem to be in accordance with the results from~\cite{guo2017}. All of the models are somewhat miscalibrated. Some of the Bayesian approaches, however---in particular the models based on MC-Dropout and SWAG---are better calibrated  than their point estimate DNN counterparts.

\begin{table}[ht!]
    \centering
    \caption{Expected calibration errors (ECE) for \textit{CIFAR-10}, \textit{SVHN} and \textit{FashionMNIST} equivalently. Lower values indicate better calibrated models.}
 \begin{tabular}{lrrrrrrrr}
	\toprule
	Models & & CIFAR10 & & & SVHN & & & FashionMNIST \\
	\midrule
	VGG16-SGD & & 0.0677236 & & & 0.0307552 & & & 0.032874 \\
	VGG16-MC Dropout & & 0.0423307 & & & 0.0155248 & & & 0.015468 \\
	VGG16-SWAG & & 0.0499478 & & & 0.0204564 & & & 0.017393 \\
	PreResNet164 & & 0.0309783 & & & 0.0238022 & & & 0.021633 \\
	PreResNet164-MC Dropout & & 0.0338277 & & & 0.0161251 & & & 0.020146 \\
	PreResNet164-SWAG & & 0.049338 & & & 0.0089618 & & & 0.015430 \\
	WideResNet28x10 & & 0.0200257 & & & 0.0210420 & & & 0.021181 \\
	WideResNet28x10-MC Dropout & & 0.0255283 & & & 0.0147181 & & & 0.020818 \\
	WideResNet28x10-Swag & & 0.0097967 & & & 0.0082371 & & & 0.011501 \\
	DeepGaussProcess & & 0.1418236 & & & 0.3275412 & & & N/A \\
	\bottomrule
\end{tabular}
    \label{tab:ece}
\end{table}

Notice that all models exhibit high accuracy on the final test set (shown in  Table~\ref{tab:accuracy}). This illustrates that a model can be very accurate but miscalibrated, or equivalently a model can be very well calibrated but inaccurate. There is no real correlation between between calibration and accuracy of a model. It is also known~\cite{guo2017} that as the complexity of the model increases the calibration error increases as well.

\begin{table}[ht!]
 \caption{Accuracy of all the models on both datasets \textit{CIFAR-10}, \textit{SVHN} and \textit{FashionMNIST}.}
    \centering    
    \begin{tabular}{lrrr}
    \toprule
    Models & CIFAR10 & SVHN & FashionMNIST \\
    \midrule
    VGG16-SGD & 94.40 & 97.10 & 95.76 \\
    VGG16-MC Dropout & 93.26 & 96.87 &  95.60 \\
    VGG16-SWAG & 93.80 & 96.83 & 96.30 \\
    PreResNet164 & 93.56 & 97.90 & 96.86 \\
    PreResNet164-MC Dropout & 94.68 & 97.73 & 97.12 \\
    PreResNet164-SWAG & 93.14 & 97.69 & 97.18 \\
    WideResNet28x10 & 94.04 & 97.44 & 97.16 \\
    WideResNet28x10-MC Dropout & 95.54 & 97.63 & 97.08 \\
    WideResNet28x10-SWAG & 95.12 & 97.95 & 97.18 \\
    DeepGaussProcess & 91.00 & 93.00 & N/A \\
    \bottomrule
    \end{tabular}
    \label{tab:accuracy}
\end{table}

\begin{figure}[ht!]
    \centering
    \includegraphics[width=5in]{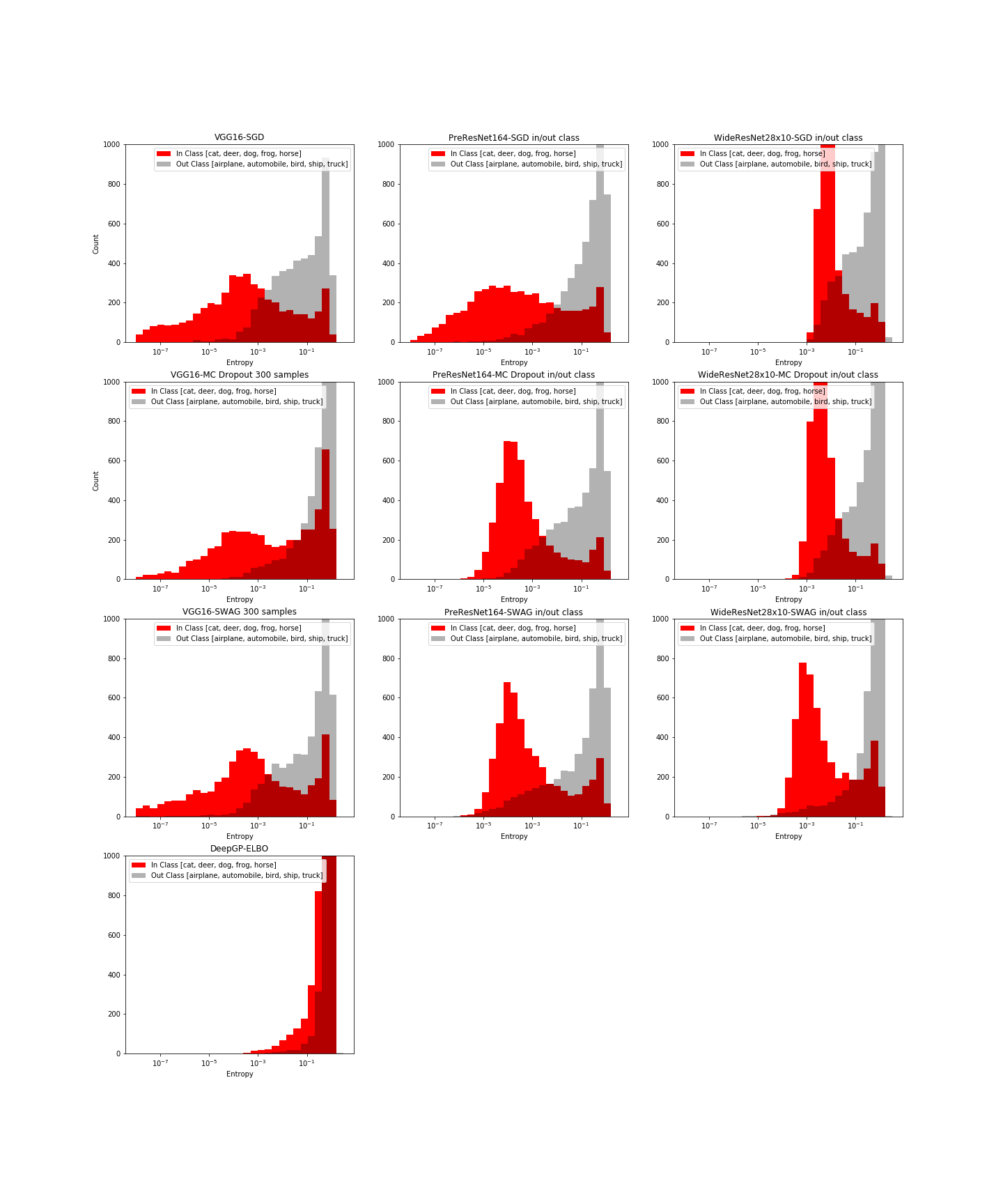}
    \caption{Out of sample distributional entropy plots for all models on CIFAR-10 (5 + 5) categories.}
    \label{fig:entropy_cifar}
\end{figure}

\begin{figure}[ht!]
    \centering
    \includegraphics[width=5in]{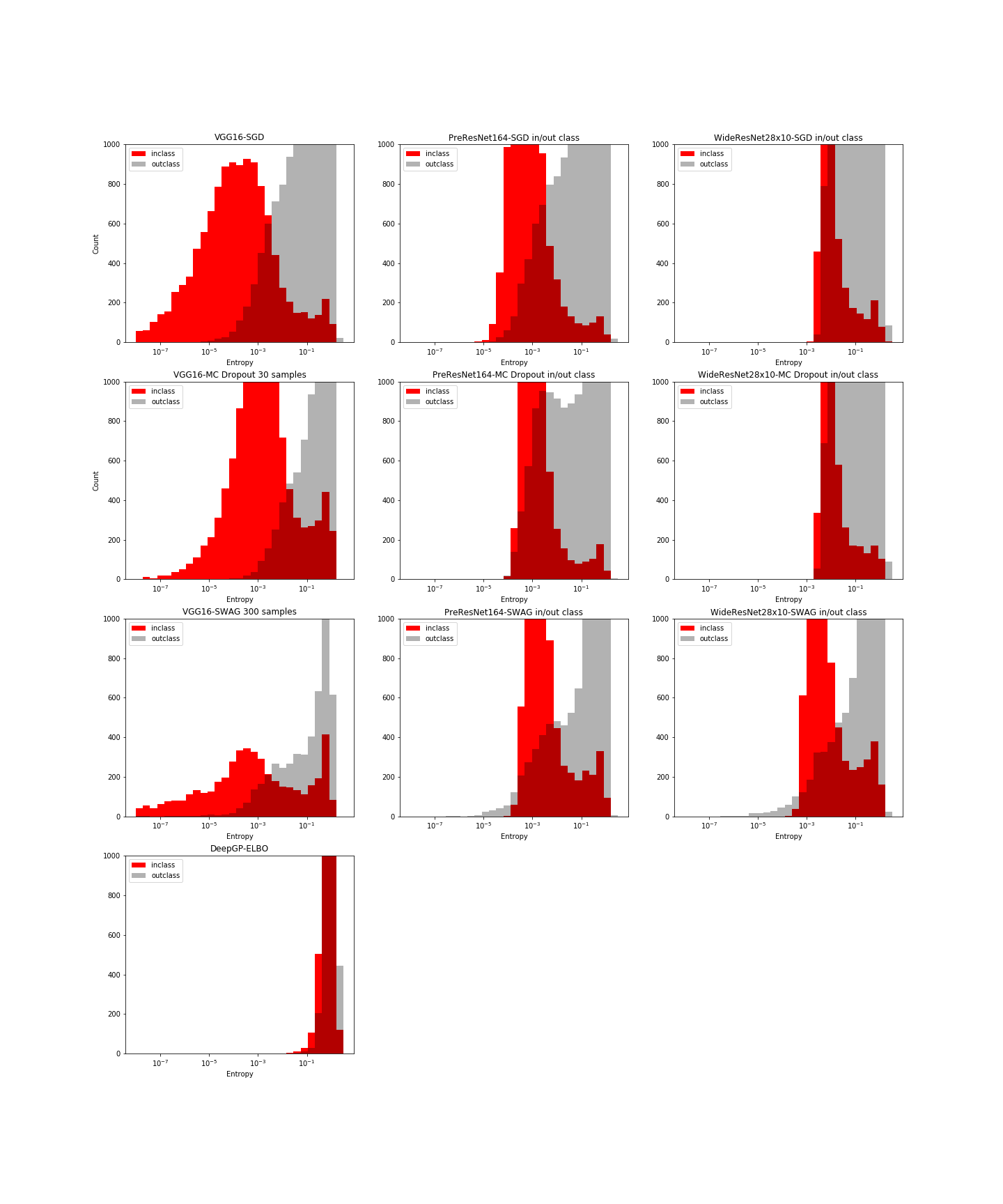}
    \caption{Out of sample distributional entropy plots for all models on SVHN (5 + 5) categories.}
    \label{fig:entropy_svhn}
\end{figure}

\begin{figure}[ht!]
    \centering
    \includegraphics[width=5in]{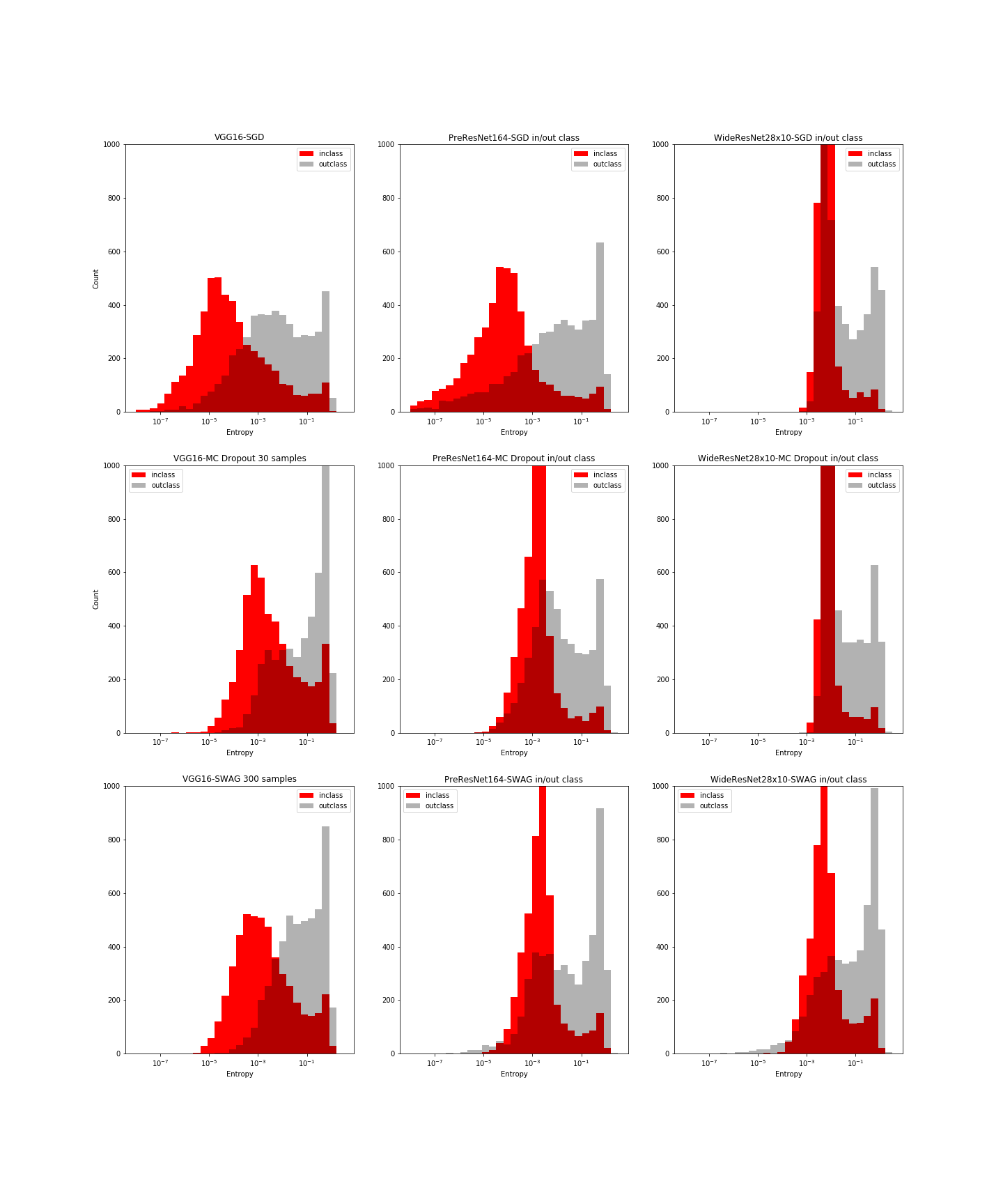}
    \caption{Out of sample distributional entropy plots for all models on FashionMNIST (5 + 5) categories.}
    \label{fig:entropy_fmnist}
\end{figure}

In order to evaluate and demonstrate the ability of the models to handle out of sample instances we divided each of the \textit{CIFAR-10} and \textit{SVHN} datasets into two halves containing 5 categories each. These partitions represent in and out of distribution samples. In the discussion that follows this is indicated with the parenthesis (5 + 5) next to the dataset name to denote that the model was trained on only 5 categories representing in distribution samples and at test time it was evaluated on the other 5 categories to simulate out of sample instances. The results are illustrated in Figures~\ref{fig:entropy_cifar},~\ref{fig:entropy_svhn} and~\ref{fig:entropy_fmnist}
for \textit{CIFAR-10}, \textit{SVHN} and \textit{FashionMNIST} respectively, and summarised in Table~\ref{tab:divergence}. Table~\ref{tab:divergence} provides a summary of Figures~\ref{fig:entropy_cifar},~\ref{fig:entropy_svhn} annd~\ref{fig:entropy_fmnist},
by measuring the  symmetric KL divergence, between the distribution of class confidence entropies of each model for the in and out of sample instances.
\begin{table}[ht!]
    \centering
    \caption{Symmetric $D_{KL}$ divergence between in and out of distribution splits of \textit{CIFAR-10} (5 + 5) and \textit{SVHN} (5 + 5). Higher values indicate the ability of a model to flag out of sample instances with high uncertainty.}    
     \begin{tabular}{llrlrrrr}
         \toprule
		    Models & & CIFAR10 & & &  SVHN & & FahsionMNIST \\
		    \midrule
		    VGG16-SGD & &  2.952740 & & & 5.638210 & & 1.519889 \\
		    VGG16-MC Dropout &  & 3.849440 &  &  & 6.273212 & & 1.405272 \\
		    VGG16-SWAG &  & 2.375000 & & &  5.058158 & & 2.060559 \\
		    PreResNet164 &  & 4.244287 & & & 3.375254 & & 1.627958 \\
		    PreResNet164-MC Dropout &  & 2.879347 & & & 2.705263 & & 1.195753 \\
		    PreResNet164-SWAG &  & 1.810131 & & & 3.344560 & & 1.496679 \\
		    WideResNet28x10 & & 2.181033 & & & 3.051318 & & 1.199489 \\
		    WideResNet28x10-MC Dropout &  & 2.929135 & & &  2.995543 & & 1.208472  \\
		    WideResNet28x10-SWAG &  & 2.780160 & & & 3.646878 & & 1.667415 \\
		    DeepGaussProcess & & 0.801246 & & & 0.153648 & & N/A \\
         \bottomrule
   \end{tabular}
    \label{tab:divergence}
\end{table}

Together these results suggest that the Bayesian methods are better at identifying out of sample instances. Although the result is not clear cut, in some cases the point estimate networks get higher divergence scores than the Bayesian ones, overall the results point in the Bayesian direction.

\section{Conclusion}
\label{sec:conclusion}
In conclusion, as we have showed that point estimate deep neural networks indeed suffer from poor calibration and inability to identify out sample instances with high uncertainty. Bayesian deep neural networks provide a principled and viable alternative that allows the models to be informed about the uncertainty in their parameters and at the same time exhibits a lower degree of sensitivity against noisy samples compared to their point estimate DNN. This suggests that this is a promising research direction for improving the performance of deep neural networks.

 \vspace{3mm}
 \noindent\textbf{Acknowledgement.} This   work   was   supported   by   Science   Foundation   Ireland   under   Grant   No.15/CDA/3520 and Grant No. 12/RC/2289.

\clearpage
\bibliographystyle{splncs03}
\bibliography{aics-sample}

\clearpage
\renewcommand{\tablename}{Figure}
\begin{table}
\begin{tabular}{ll}
\multicolumn{2}{c}{\includegraphics[width=2in]{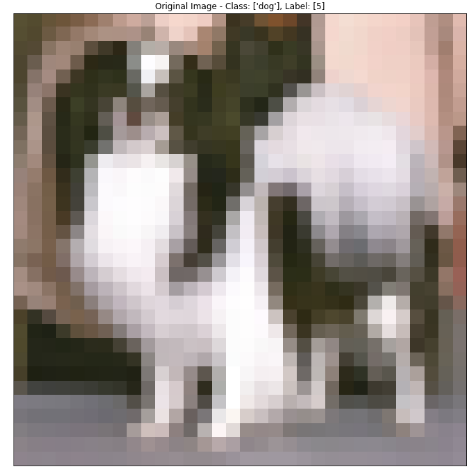}} \\
\includegraphics[width=3in]{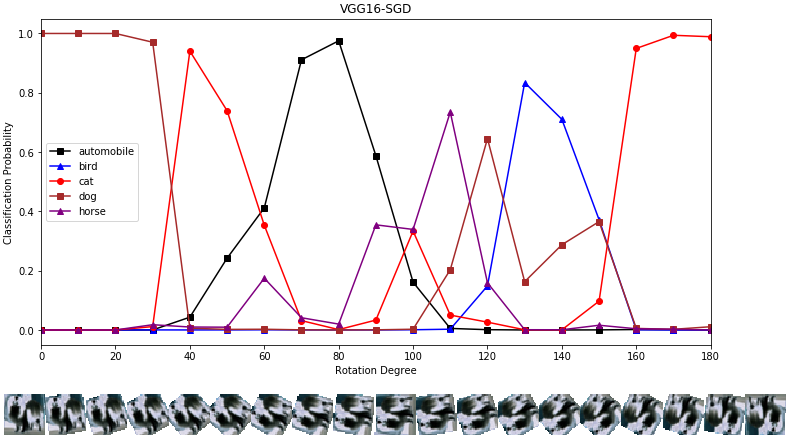} &  \includegraphics[width=3in]{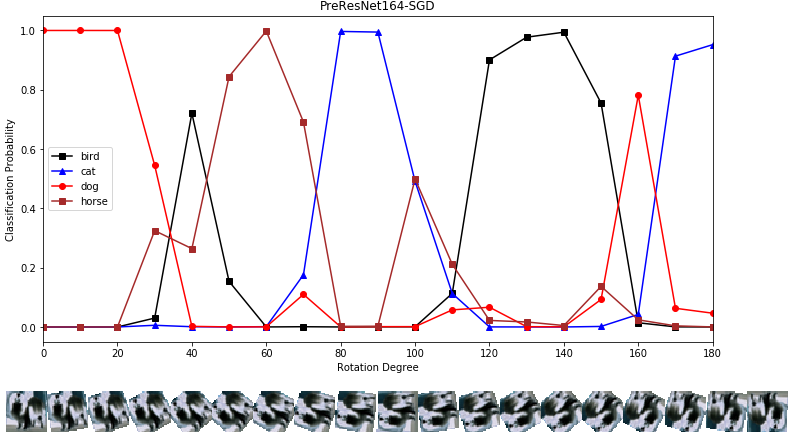} \\
\includegraphics[width=3in]{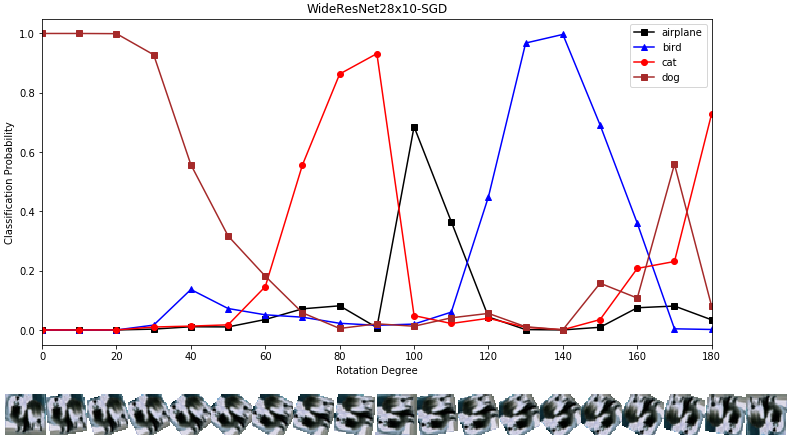} & \includegraphics[width=3in]{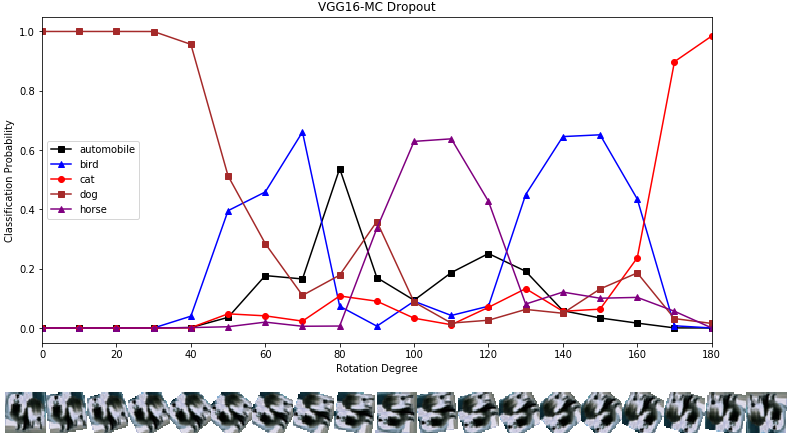} \\
\includegraphics[width=3in]{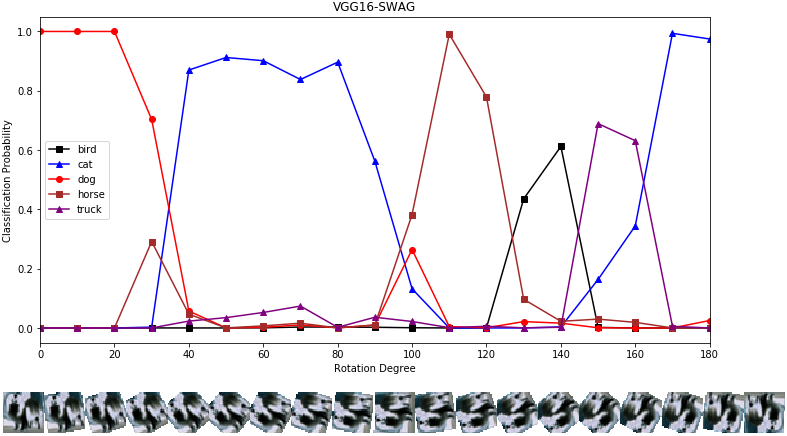} & \includegraphics[width=3in]{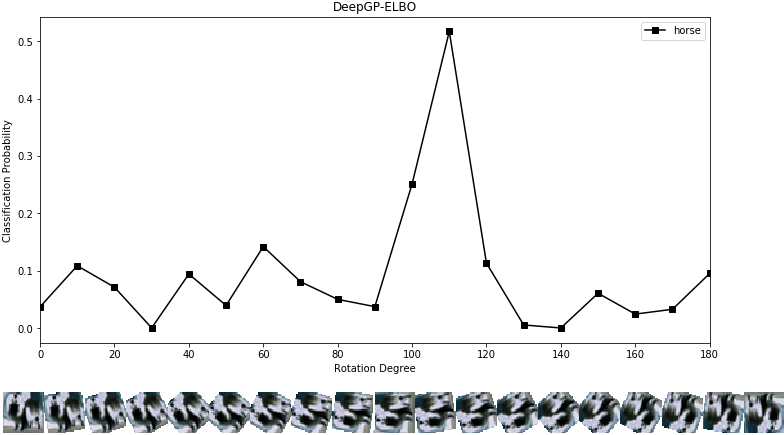} \\
\end{tabular}
\caption{Evaluating models on out of distribution samples, simulating the input query image (i.e.~dog) as OoD by rotating it with $30^{o}$ degree angle and recording the models' predictions.}
\end{table}

\end{document}